\documentclass[a4paper]{article}

\usepackage[english]{babel}
\usepackage[utf8]{inputenc}
\usepackage{amsmath}
\usepackage{graphicx}
\usepackage{booktabs}
\usepackage[colorinlistoftodos]{todonotes}
\usepackage{biblatex}
\usepackage{algpseudocode}
\usepackage{bm}

\title{A Minimal Active Inference Agent}

\author{Simon McGregor$^1$, Manuel Baltieri$^1$ and Christopher L. Buckley$^1$
\\\mbox{}
\\$^1$University of Sussex, Brighton, UK \\
s.mcgregor@sussex.ac.uk, m.baltieri@sussex.ac.uk, c.l.buckley@sussex.ac.uk}

\date{\today}

\bibliography{sources}

\begin{document}
\maketitle

\begin{abstract}
Research on the so-called ``free-energy principle'' (FEP) in cognitive neuroscience is becoming increasingly high-profile. To date, introductions to this theory have proved difficult for many readers to follow, but it depends mainly upon two relatively simple ideas: firstly that normative or teleological values can be expressed as probability distributions (active inference), and secondly that approximate Bayesian reasoning can be effectively performed by gradient descent on model parameters (the free-energy principle). The notion of active inference is of great interest for a number of disciplines including cognitive science and artificial intelligence, as well as cognitive neuroscience, and deserves to be more widely known. 

This paper attempts to provide an accessible introduction to active inference and informational free-energy, for readers from a range of scientific backgrounds.  In this work introduce an agent-based model with an agent trying to make predictions about its position in a one-dimensional discretized world using methods from the FEP.
\end{abstract}

\section{Introduction}
Active inference is the name given by Karl Friston to a process of an agent’s simultaneously co-constrained inference regarding the world combined and directed actions to change it, understood specifically as the minimisation of an information-theoretic quantity describing the instantaneous relation of a system to its “typical” sensory environment.

According to proponents of active inference, all mechanical implementation of cognitive behaviour (including both the active and inferential aspects of cognition) can be reduced to the minimisation of a single informational “free energy” quantity; moreover, this quantity can be assigned an interpretation using cognitive concepts resembling beliefs and intentions.

This theory has attracted a great deal of attention (both positive and negative), since it claims to unify a wide variety of principles in cognitive science, machine learning and neuroscience. However, a clear and balanced discussions of the framework's strengths and weaknesses is still lacking. 

We suspect that one of the barriers preventing a wider understanding of the theory is the language which been used to describe it. The language is technically dense, using terms like 'surprise', 'generative', 'ensemble', 'variational', and can seem to beg philosophical questions in a way which is not strictly mandated by the formalism of the active inference theory. For instance, the framework is frequently characterised as claiming that cognitive agents act so as to ``confirm their predictions'', or ``minimise their surprise''; we will see that alternative interpretations exist for the same formal assertions. 

Furthermore one of the central strengths of the formalism is its potential to unite an understanding off   action and perception within a single  unified framework \cite{Friston10}. However understanding the  exact implications of this will likely require the utilisation of   agent based approaches which to  date  have largely been absent.

This paper aims to clarify the foundations of the active inference or ``free-energy'' framework in terms which can be understood by a wide audience.  We start by giving a brief conceptual overview of the information Free-Energy (IFE)  principle and active inference in an intuitive language

We ground our discussion by developing a  Free-Energy formalism for simple active agent behaving in  one-dimensional discrete-time world. This agent based approach allows us to  examine the relationship between action and perception that emerges from the interaction of the agents' internal  model and the world around it.  We finish by the outlining the main contributions and the known current limitations of the IFE . In particular we  will address some of the claims made regarding free-energy or active inference, and attempt to deflate them in a way which may help readers make more sense of the underlying ideas. 
 
%
%



The foundational premise of the IFE  principle is that adaptive agents are defined by their ability occupy only a limited repertoire of physical states  \cite{Friston10}. Adaptive agents  achieve this by exporting thermodynamic entropy (thus resisting the third law) and preserving the defining traits that comprise their very identity. However agents do not have  direct access to states that constitute there world and body and instead must work vicariously on sensory data do this. Formally speaking, the active inference framework proposes that agents do this by acting to minimise an information-theoretic quantity derived from sensory data known as surprisal. The interpretation of this quantity is somewhat subtle, and it should absolutely not be conflated with the psychological phenomenon we call surprise.

Friston has previously asserted that agents which export thermodynamic entropy must do so by minimising the informational entropy of their sensory inputs (considered as an empirical distribution sampled over time). He argues that the only tractable way of doing so is to minimise informational free energy. Thus, the very fact that a living system maintains its organisation in the face of environmental perturbation supposedly justifies adverting to the ``free-energy principle'' \cite{Friston2006free}.

One of the most interesting aspects of the active inference framework is that two things which we generally regard as quite distinct are given the same formal treatment: on the one hand, beliefs or predictions about sensory input and the external world which underlies it, and on the other hand intentions regarding the outcomes of our actions. In this sense, active inference blurs the distinction between a prediction and an intention; both are treated as conceptions-of-the-future, with the main difference being in what physical variables change most when the reality turns out to be different from the expectation \cite{Friston10}.

Although the free-energy literature uses the terms ``prediction'' and ``expectation'' interchangeably for this concept, it seems that only the verb ``expect'' in English really captures the ambiguity between intention and prediction which the active inference framework rests upon. It is unfortunate that the word ``expectation'' also has a technical meaning which is relevant to active inference, i.e. the mean of a probability distribution. Hence, we propose the term \emph{``expectance''} to refer to the superclass of probabilistic beliefs and desires implicit in the active inference framework.

In this language, active inference asserts that agents constantly behave so as to reduce the discrepancy between their expectances and reality; the resolution of discrepancy by changing internal variables is known as “updating one’s predictions”, while the resolution of discrepancy by changing reality is known as “acting on one’s environment”.

Crucially, under active inference both of those dynamics are always occurring simultaneously. It is not necessary to pre-judge which of those dynamics will predominate by calling the expectance a “prediction” or an “intention” in advance of observing the outcome. 

The typical presentation of active inference states that agents behave so as to make their “predictions” come true. We believe this wording erases the distinction between internal and external resolution of expectance violation, in an unhelpful way. Intuitively, if an agent “predicts” its arm to move to the left, and the arm instead stays in place, we would expect it to change internally so as to make different predictions; no action on the world would be presupposed by the discrepancy. On the other hand, if the agent “intends” its arm to move to the left, and the arm instead stays in place, we would expect it to exert a greater force on its arm by contracting the relevant muscles. 

Of course, it is entirely possible that the agent could begin by attempting to move its arm and, in doing so, learn that its range of movement is restricted by external forces. The expectance violation can be resolved by coming to a halfway point in which the agent “updates its predictions” to reflect the fact that its movement is restricted, “modifies its intentions” to be consistent with what is achievable, and “acts on the world” by moving within its available range. Such an equilibrium can emerge naturally from a continuous-time process where informational free energy is minimised along distinct internal dimensions simultaneously.

Recent discussions of active inference have tended to occur within the context of a neuroscientific theory known as the predictive coding hypothesis \cite{clark:next}. The predictive coding idea is essentially this: the main function of sensory neurons is to predict their sensory inputs (using efferent signals from higher brain regions) and report their prediction errors, with the consequence that afferent sensory signals are nonzero only when they carry new information. The hierarchical predictive coding hypothesis additionally posits that this mode of operation extends to multiple layers of neurons, with each layer providing input for the next.

Friston states that, if one makes a number of (non-trivial) mathematical assumptions and approximations, active inference via informational free-energy minimisation can be implemented in a hierarchical predictive coding neural model. Within this hierarchical predictive coding model for active inference \cite{friston08a}, efferent (motor) signals are treated similarly to other signals. The workings of this process can be a little difficult to understand at first; the idea is that the neural architecture imposes certain specific patterns of expectances on instantaneous kinesthetic sensations \cite{Friston10}. These expectances correspond to the agent’s desired behaviour, such that the expectancy error directly becomes a motor signal.

Unfortunately, it is not at all clear that a direct mapping between expectancy error on kinesthetic sensation and motor signal makes sense within a general framework of cognition. Firstly, the brain’s kinesthetic model is learned, rather than genetically determined, and consequently its relationships with motor signals need to be learned as well. For active inference to work as a universal explanatory principle, the formalism needs to be able to explain how this mapping arises in the first place; an isomorphism between the semantic meaning of motor signals and the semantic meaning of kinesthetic sensations cannot simply be assumed.

Secondly, general theories of cognition ought to be widely relevant; their application cannot be restricted to humans, mammals or even animals. The principles which underlie adaptive behaviour extend also to plants, single-celled organisms, and arguably even to systems such as social insect colonies or non-biological dissipative structures. How does one model the error in kinesthetic sensory expectance of a bacterium? Or a swarm of bees? This problem becomes most evident when one attempts to construct a minimal simulation based on active inference. 

Fortunately, the issue is caused by not intrinsic to the active inference framework itself, but only by the simplification which treats kinesthetic sensation as a special mode of sensation. If this assumption is not made, the free-energy equations give a general solution: filter the choice of action through the agent’s (current provisional) model of action’s effect on the whole of its sensorium. We show in this paper that such an approach works perfectly well for a simple simulated artificial agent.

\section{The Free Energy Formalism}
\label{sec:mat}
%
%

\begin{table}[h]
	\begin{center}
		\begin{tabular}{cl}
			\toprule
				State variables					& Meaning\\
			\midrule
				\textbf{$\Psi_i, \psi_i$}		& World state at time $t_i$\\
                \textbf{$B_i, b_i$}				& Internal (brain) state at time $t_i$\\
				\textbf{$S_i, s_i$}				& Sensory input at time $t_i$\\
				\textbf{$A_i, a_i$}				& Motor trajectory between $t_i$ and $t_{i+1}$\\
			\bottomrule
		\end{tabular}
		\caption{State variables}
		\label{tab:statevar}
	\end{center}
\end{table}

Following the work by Dayan et al. \cite{dayan:hm} on the Helmholtz machine, we define 2 densities that will constitute the free energy term to be optimised:
\begin{itemize}
	\item a generative density $p(\psi, s \mid m)$ representing the joint probability of world states $\psi$ and sensory input $s$ based on a probabilistic predictive model $m$ by the agent or brain
	\item a recognition density $q(\psi \mid b)$ encoding the agent's (or brain's) beliefs about the causes $\psi$ with a set of brain states $b$ that fully describe these beliefs.
\end{itemize} 

IFE is then defined as:

\begin{equation}
\label{fe}
	F(s, b)  = \int_{\psi} q(\psi \mid b) \ln \frac{q(\psi \mid b)}{p(\psi, s \mid m)} \, d\psi\\
\end{equation}

Optimising this term requires then a brain (agent) to be able to
\begin{itemize}
	\item change its \textbf{perception}, using the internal states $b$ to alter its beliefs $q(\psi \mid b)$ thus enhancing its ability of explaining the world causes as they are produced by the generative model, or
	\item directing its \textbf{action}s to result in different sensory input in the generative density $p(\psi, s \mid m)$ that match the agent's beliefs.
\end{itemize}

Two alternative analytical forms for the IFE better show dependence on perception and action respectively, helping us to understand the contributions to the free energy minimisation of both:

\subsection{Perception}

\begin{equation}
	\label{fe_2}
	F(s, b) = D ( q(\psi \mid b) \mid \mid p(\psi \mid s, m)) - \ln p(s \mid m)
\end{equation}

where $D ( q(\psi \mid b) \mid \mid p(\psi \mid s, m ))$ is the Kullback-Leibler (KL) divergence (defined in the appendix \ref{appx:kldiv}), between the recognition density ($q(\psi \mid b)$) and the true posterior of the world states ($p(\psi \mid s$); $- \ln p(s \mid m)$ is the {\em surprise} about the sensory input the brain cannot directly evaluate. Changing the set of brain states $b$ allows to approximate the posterior of the world states with the brain's beliefs about the world. In the ideal case where the two coincide, free energy would be equal to the surprise (the KL divergence would be zero, appendix[\ref{appx:kldiv}]) thus being able to interpret this term otherwise hidden to the brain.

The optimal internal state $b^{opt}$ is defined as:
\[ b^{opt} = \arg \min_{b} F(s, b) \]

\subsection{Action}
An alternative form shows instead how the free energy depends on sensory input, that can be thought as dependent on a $a$ which represent the set of actions a system can perform in a certain environment, giving then s$(a)$.
\begin{equation}
	\label{fe_3}
	F(s, b) =  D ( q(\psi \mid b) \mid \mid p(\psi \mid m )) -\langle \ln p(s(a) \mid \psi, m) \rangle_{q}
\end{equation}

with $D ( q(\psi \mid b) \mid \mid p(\psi \mid m ))$, KL divergence between the recognition density $q(\psi \mid b)$ (i.e. posterior belief about the causes/world states) and the prior (belief) $p(\psi \mid m ))$ of the world states $\psi$; $\langle \ln p(s(a) \mid \psi, m) \rangle_{q}$ being the expectation about sensations s$(a)$ under the density $q$. In this formulation we can see how free energy is minimised using an optimum action $a^{opt}$, that will sample inputs as predicted by the recognition density.
\[ a^{opt} = \arg \min_{a} F(\text{\~{s}}(a), b) \]

\section{A discretized approach to the FEP}

%
%

In this work we will consider a discrete representation of our state variables, defining specifically $\Psi', \psi'$ and $B', b'$ respectively as world and internal states at time $t_{i+1}$.

In its discrete version, Informational Free Energy (IFE) becomes a quantity relating an action $a$ and a sensation $s$ to two possible internal states $b$ and $b'$ based on a generative and a recognition densities which are attributed to the agent by the theorist.

\begin{table}[h]
	\begin{center}
		\begin{tabular}{cl}
			\toprule
				Densities						& Name\\
			\midrule
				\textbf{$q(\Psi' \mid B') = P(\Psi' \mid B')$}				& Recognition density\\
                \textbf{$p(\Psi', S \mid B, A) = P(\Psi', S \mid B, A)$}	& Generative density\\

			\bottomrule
		\end{tabular}
		\caption{Key probability densities }
		\label{tab:densities}
	\end{center}
\end{table}

The recognition density described the agent's internal ``encoding'' of external world states, while the generative density represents the agent's ``predictive model'' of physical dynamics. 

Most active inference papers use a continuous-space definition of the informational free energy $F$\cite{friston:reinf, Friston2006free, Friston10}, but in this paper we will consider a discrete-space version:
\begin{equation}
	F(b', b, s, a) = \sum_{\psi'} q(\psi' \mid b') \ln \frac{q(\psi' \mid b')}{p(\psi', s \mid b, a)}
\end{equation}

IFE has several interesting properties, corresponding to different rearrangements of the formula, which give rise to different interpretations \cite{Friston10} as we described in the previous section. We will focus in this section on its role in approximate Bayesian filtering (i.e. ongoing inference about a changing world state). Following our discretized definition, the equation can be rewritten as
\begin{equation}
	F(b', b, s, a) = D_{KL} ( q(\psi' \mid b') \mid \mid p(\psi' \mid s, b, a)) - \ln p(s \mid b, a)
\end{equation}
where the first term is the KL divergence between $q(\psi' \mid b')$ and $p(\psi' \mid s, b, a)$, as defined previously. The second one is the informational surprisal about the sensory inputs $s$ that cannot be directly evaluated by the agent. Since the KL divergence is always a positive quantity, free energy represent an upper bound of surprisal, meaning that by minimising IFE we indirectly optimise surprisal at the same time.

%
%
Assuming the agent does not expect sensation $s_i$ at time $t_i$ to depend on action $a_i$ that originates a movement from time $t_i$ to time $t_{t+1}$ , i.e. $p(s \mid b, a) = p(s \mid b)$, the second term is independent of $b'$ and $a$. This has two interesting consequences:

\begin{enumerate}
\item {\emph{Approximate Inference} {Minimising $F(b_1, b_0, s_0, a_0)$ with respect to $b_1$ corresponds to approximate Bayesian inference regarding the state of the world at time $t_1$.}}

\item {\emph{Optimal Control} {Minimising $F(b^*, b_0, s_0, a_0)$ with respect to $a_0$ corresponds to optimal control (subject to the agent's assumptions about world dynamics) if $b^*$ encodes a desired distribution over world states at time $t_1$.} }
\end{enumerate}

Note that the second statement draws a distinction between belief $P(\psi_1 \mid b_1)$ and desire $P(\psi^* \mid b^*)$ which does not reflect the original maths of the active inference framework \cite{Friston09}. This will be helpful for expository purposes. Mechanisms by which goal-directed action can be encoded in the pure active inference formalism, without distinguishing predictions from intentions, will be discussed later.

We have not mentioned the actual dynamics of the environment; in fact, they are not necessary for this analysis. We assume that the only interface between the agent and its environment is the agent's sensorimotor dynamics; hence, we are free to consider how the agent's (attributed) cognitive dynamics proceed when it is fed an arbitrary series of sensory inputs by an experimenter resembling Descartes' ``malicious demon''. In either case, we expect that the agent's behaviour should be approximately rational, given the beliefs and intentions we have attributed it. 

\subsection{Practical IFE Minimisation}
The value of IFE minimisation as a computational model of active inference, either for machine learning purposes or in neuroscience modelling, depends on how realistic proposed mechanisms for minimising IFE are. In the general case, IFE minimisation is not much more computationally tractable than the sum required to perform exact Bayesian inference, because it involves a sum (or, in the continuous case, integral) over all possible world states. It is of course conceivable that the brain can implicitly compute this integral through physical means which are expensive to simulate computationally. However, IFE is significantly easier to compute when certain certain conditions are met regarding how the agent-environment dynamics work and how the agent represents world states. These conditions, along with additional approximations, allow IFE minimisation to form a tractable computational approach to active inference \cite{friston_07, friston08a} and are argued \cite{Friston10} to provide an elegant model of neural function.

\subsection{Unifying Beliefs and Desires}
We have indicated why minimising $F(b_1, b_0, s_0, a_0)$ with respect to $b_1$ constitutes approximate Bayesian filtering, and why minimising $F(b^*, b_0, s_0, a_0)$ with respect to $a_0$ constitutes optimal control. In a machine learning context, it will frequently be convenient to maintain a distinction between belief and intention, and the computational cost of maintaining the distinction is small, because the function $F$ (and its derivatives) can be called with arbitrary arguments.

However, the free-energy formulation in neuroscience postulates a different mechanism. The \emph{same} IFE term is used to provide a gradient term which is applied to the dynamics of both internal state and action \cite{Friston10}. Arguably, this is more biologically plausible since an agent has only one brain state, but the interpretation is less clear.

\section{A Minimal Free-Energy Agent}
This section outlines in detail how the active inference formalism works for a ``minimal'' abstract model of active agency. Agents are modelled as organisms that inhabit a single discrete space (a ``cell'') in a one-dimensional discrete-time world, and which are sensitive to a chemical which occurs in their environment. The world has periodic boundary conditions (i.e. it ``wraps around'' at the edges), and each cell contains a concentration of the chemical. The concentration of the chemical follows a gradient across the environment, being highest in a ``source'' cell and lowest in the cell furthest from the source; if the reader likes, this can be imagined as the result of a diffusion process.

The agent's interactions with its environment are extremely low-bandwidth: it has a 1-bit sensor, which fires with a probability proportional to the chemical's concentration, and a 1-bit motor, which attempts to move it one way or the other along the 1d world. We will arbitrarily model the agent as having a ``desire'' to move to a particular spot in its world (relative to the ``source'').

This system is simple enough that exact Bayesian inference can be performed directly, making approximate schemes such as free-energy minimisation unnecessary. Hence, free-energy minimisation for this system is done purely for didactic purposes, with the advantage that the results can be compared to exact Bayesian posteriors. For more complex systems, exact Bayesian inference rapidly becomes intractable, and approximate methods are necessary.

\begin{figure}[!htb]
\centering
\includegraphics[scale=.75]{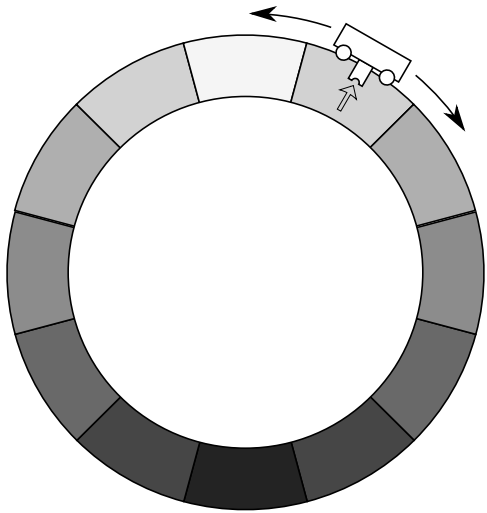}
\caption{Illustration of agent-environment system. The agent has a sensor which reads $High$ or $Low$ and is sensitive to chemical concentration. The agent's motor can attempt to move the agent clockwise or anticlockwise.}
\label{fig:digraph}
\end{figure}

\subsection{Definitions}

We will begin by describing the simulation framework for our agents and the environments they inhabit. The agent-environment dynamics are modelled in discrete time steps. The agent is represented as a system with an internal `brain' state $b$. After emitting an action $a$, the agent receives a sensory input $s$ and updates its brain to a new state $b'$ based on $b$, $a$ and $s$. It then emits a new action $a'$ based on $b'$, and the cycle starts again.

It will sometimes be worth considering how the agent's dynamics proceed when it is fed an arbitrary series of sensory inputs by an experimenter resembling Descartes' ``malicious demon'', but in general we are interested in the coupled dynamics of an agent and its ``natural'' environment. The environment will be represented as a system with a state $\psi$, which changes to state $\psi'$ depending on its current state and the agent's action, and the sensory input of the agent depends on the instantaneous state of the environment.
%
%


This is a fairly standard approach to modelling discrete-time agency, but in fact there is nothing ``agentive'' in the equations - they describe an arbitrary coupled dynamical system. As cognitive scientists we will be assigning a \emph{cognitive interpretation} to the agent and its dynamics; this interpretation can (and should) be distinguished from the bare `physical' dynamics of the agent system. 

In this model, a key part of the interpretation will be the assertion that each brain state `encodes' a \emph{belief} about the world\footnote{In principle, we should distinguish the ``agent's world'', about which it can have beliefs, from the ``theorist's world'' that corresponds to its `actual' environment. For expository purposes we will ignore this distinction. }. Following Bayesian principles, the belief will take a probabilistic form: instead of holding only clear-cut beliefs, the agent's brain will encode beliefs that can have varying degrees of uncertainty. It is important to note that these probabilities are understood as degrees of \emph{subjective uncertainty} for the agent, not as the `objective' probability of any simulated outcome. 

The agent-environment system's `physical' dynamics can be directly observed over time, producing a series of brain states. The theorist can then project these brain states onto a corresponding series of Bayesian beliefs: a ``belief dynamics'' of the agent system under a particular cognitive interpretation.

\subsection{Formulation}
In our case, the environment is modelled as having a length of $n=16$ cells and its state $\psi$ consists only of the agent's position $x$ ($\psi = x$). The agent's brain state $b$ will therefore encode a probability distribution $P(\psi \mid b)$ over possible cell locations.

The recognition density, which encodes the agent's internal beliefs about his position at time $t+1$ will be described as follows. The agent's brain state $b'$ is a vector of real numbers $b'_1, \cdots, b'_n$ and we use a softmax encoding:
\begin{equation}
	q(\psi' \mid b') = P(\psi' \mid b') = \frac{e^{b'_\psi}}{\displaystyle \sum_{i} e^{b'_i}}
\end{equation}
	As described the section above, for an active inference agent we define a sort of target brain state: the state the agent would ``prefer'' its brain to be in, subject to the constraint that its brain is updated rationally. Another interpretation is that it encodes the location distribution which the agent would prefer to occupy over an ensemble of possible Universes. Consequently we write an agents \textbf{intention} as  
\begin{equation}
	q(\psi^* \mid b^*) = P(\psi^* \mid b^*) = \frac{e^{b^*_\psi}}{\displaystyle \sum_{i} e^{b^*_i}}
\end{equation}

The generative density associated with the agent's predictions about his next position $\psi'$ given his actual state, can be decomposed as:
\begin{equation}
	p(\psi', s \mid b, a) = P(\psi', s \mid b, a) = \sum_{\psi \in X} P(\psi' \mid s, b, a, \psi) P(s \mid b, a, \psi) P(\psi \mid b, a)
\end{equation}
We make some more assumptions on the environment to simplify the last constructs, that will be named \textbf{environmental dynamics}, \textbf{sensory dynamics} and \textbf{brain encoding} (this last one following a specific assumption which will simplify the third term into the agent's belief at the actual time step $t$):
\begin{itemize}
	\item $P(\psi' \mid s, b, a, \psi) = P(\psi' \mid a, \psi)$, the position at time $t+1$, $\psi'$, is only influenced by the action $a$ taken from the position $\psi$ at time $t$
	\item $P(s \mid b, a, \psi) = P(s \mid \psi)$, the sensation $s$coming from the chemical source is an objective parameter, depending only on the position at time $t$ in the world
	\item $P(\psi \mid b, a) = P(\psi \mid b)$, the previous position $\psi$ at time $t$ is not influenced by the action $a$, that only affects the subsequent $\psi'$ at time $t+1$
\end{itemize}

The agents \textbf{environmental dynamics }specifies how $\psi$ (position $x$) changes as a consequence of the agent's action $a \in \{ -1, 1\}$. With fixed probability $\rho = 0.75$, the agent moves one cell in the direction determined by his action. Otherwise, the agent remains in its current cell.
\begin{equation}
P(\psi' \mid \psi, a) = 
\begin{cases}
		1 - \rho, & \text{if } \psi'=\psi \text{ mod } n\\
		\rho, & \text{if } \psi'=(\psi+a)  \text{ mod } n\\
		0, & \text{otherwise} \\
\end{cases}
\end{equation}

\textbf{Sensory Dynamics} are described by the probability of the agent's sensor registering $High$ will be set equal to the concentration of chemical in the agent's current cell $\psi$, which is assumed to fall exponentially with increasing distance from the source position $\psi_0 = n/2$, according to a decay parameter $\omega = \frac{\log 4}{16}$ and a maximum value $k=4^{\frac{-1}{16}}$:
\begin{equation}
P(s \mid \psi) = 
\begin{cases}
		ke^{-\omega |\psi-\psi_0|}, & \text{if } s = High\\
		1 - ke^{-\omega |\psi-\psi_0|}, & \text{if } s = Low
\end{cases}
\end{equation}

As explained previously, the \textbf{brain encoding }of brain state $b$ is a vector of real numbers encoded using the softmax function:
\begin{equation}
P(\psi \mid b) = \frac{e^{b_\psi}}{\displaystyle \sum_{i} e^{b_i}}
\end{equation}

Active inference simulations such as \cite{Friston09} usually assume a differential equation model in which brain state variables follow a negative IFE gradient in continuous time. In our discrete model, we use the same gradient-descent principle, but perform a number of gradient descent iterations between each simulated time step in the world.

Essentially, the simulation works as follows, after selecting a learning rate $\eta$ and number of iterations $k$ for the gradient descent:

\begin{algorithmic}
\Function{optimise}{$b, s, a$}
	\State $b' = \mathbf{0}$ \Comment{$n$-dimensional vector $0$ represents maximal ignorance}
		\For{$i \in \{ 1 \cdots k \}$}
			\State $b' \gets b' - \eta \cdot \frac{\partial}{\partial b'} F(b', b, s, a)$
			\Comment{gradient descent on $b'$}
		\EndFor
	\State \Return $b'$
\EndFunction

\\

\Procedure{simulate}{$\psi, b, b^*$}
	\Comment{initial values for $\psi, b, b^*$}
	\Loop 
		\State $s \leftarrow $ random value using $P(s \mid \psi)$
		\State $a \leftarrow \arg\min_a F(b^*, b, s, a)$ 
        \Comment{exhaustively computed}
		\State $b' \leftarrow $ \Call{optimise}{$b, s, a$}
		\State $\psi' \leftarrow$ random value using $P(\psi' \mid \psi, a)$
		\State $b, \psi \leftarrow b', \psi'$
	\EndLoop
\EndProcedure
\end{algorithmic}

\section{Results}

\begin{figure}[!htb]
\centering
\includegraphics[scale=.6]{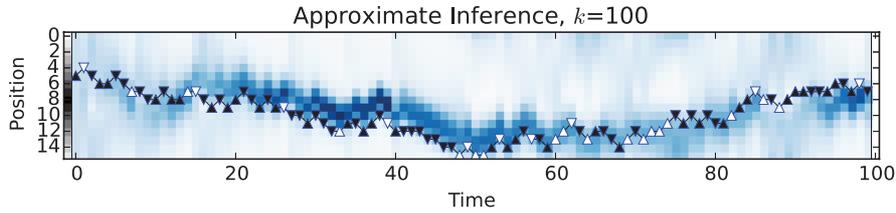}
\caption{Inferred position (shading: darker = higher credence) and actual location (triangles) of active inference agent over time. Triangle direction indicates agent's action and triangle colour indicates sensor reading (black=high; white=low). Left grey bar indicates chemical concentration gradient at different locations (darker = higher concentration). Right grey bar indicates agent's positional target (darker = more highly desired), in this case no preference is given.}
%
%
\label{fig:basicInference}
\end{figure}

An example of the dynamics of a simulated agent is given in Figure \ref{fig:basicInference}. The trajectory of the agent over time, and its subjective confidence regarding its location, can be seen in the graph.

The agent's goal is to occupy the locations towards the bottom of the graph, which are assigned a higher intentional probability than other locations (the intention distribution is shown in grey at the right-hand side of the figure). The agent's only environmental clue to its location is its sensor reading, which probabilistically detects the local concentration of a chemical. The spatial gradient of chemical concentration is shown on the left grey bar.

It can be seen that the agent is effective in simultaneous online estimation  of its location, with a brief period of confusion compensated for fairly quickly.

It is worth noting that although the ``physics'' of the agent-environment system are very simple, the agent's task is not completely trivial. The agent's target location is an arbitrary position; since the environment is symmetric, there is another location sharing the same concentration (and therefore instantaneous sensory statistics) as the target, which means that the task cannot be achieved reliably by a reactive agent. The agent therefore needs to combine simultaneous estimation and control of its position based on limited and noisy sensorimotor channels.

\begin{figure}[!htb]
\centering
\includegraphics[scale=.6]{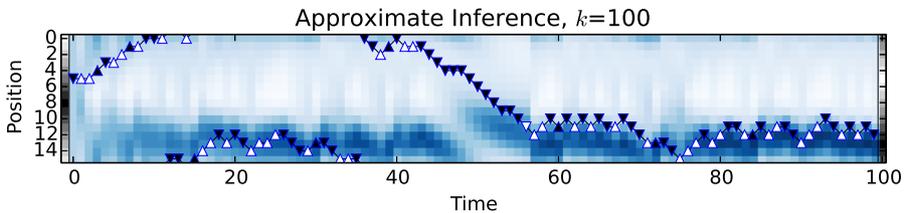}
\caption{Inferred position (shading: darker = higher credence) and actual location (triangles) of active inference agent over time. Triangle direction indicates agent's action and triangle colour indicates sensor reading (black=high; white=low). Left grey bar indicates chemical concentration gradient at different locations (darker = higher concentration). Right grey bar indicates agent's positional target (darker = more highly desired).}
\label{fig:activeInference}
\end{figure}

\begin{figure}[!htb]
\centering
\includegraphics[height=3in,natwidth=13.76in,natheight=11.91in]{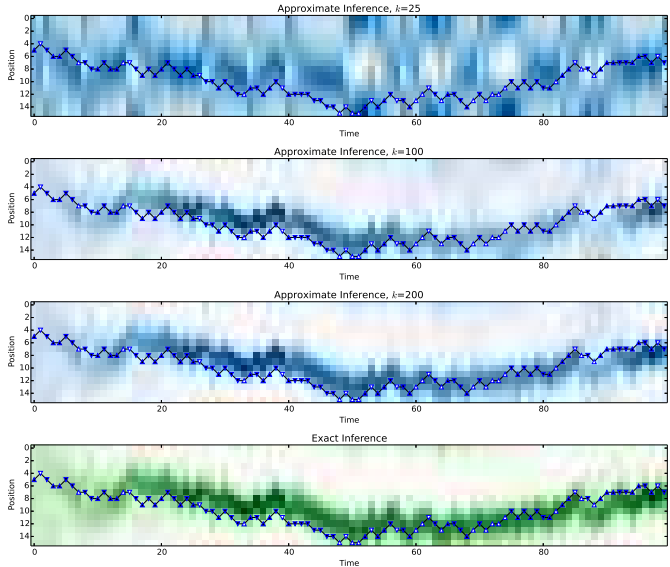}
\caption{Comparison of approximate with exact inference for the same sensorimotor data. Blue graphs, top to bottom: approximate inference with $25$, $50$ and $200$ gradient descent iterations respectively. Green graph: exact Bayesian inference. }
\label{fig:comparisonToExact}
\end{figure}

The smaller the IFE associated with the agent's internal state, the closer its inference is to the exact Bayesian posterior. We show that the degree of approximation can be controlled by varying the effectiveness of the minimisation procedure. Figure \ref{fig:comparisonToExact} shows the beliefs of different agents with increasing number of gradient descent iterations between time steps, on the same set of sensorimotor data. The exact inference is shown at the bottom for comparison. Actions were produced randomly for this figure, so the agent is effectively inferring its position over a random walk.

Gradient descent is begun from a ``null brain'' representing the uniform distribution, so with fewer gradient descent iterations the agent's inferences will tend to be biased towards higher uncertainty. We can plausibly expect information stored about past sensorimotor events to decay faster with poorer approximations to the exact posterior (which encapsulates all relevant information from arbitrary time points). Both these phenomena can be observed in the figure shown, with the upper graphs tending to be more blurred and more rapidly responsive to immediate sensory input.

\subsection{Inferential Quality and Task Performance}
\begin{figure}[!htb]
\centering
\includegraphics[scale=.4]{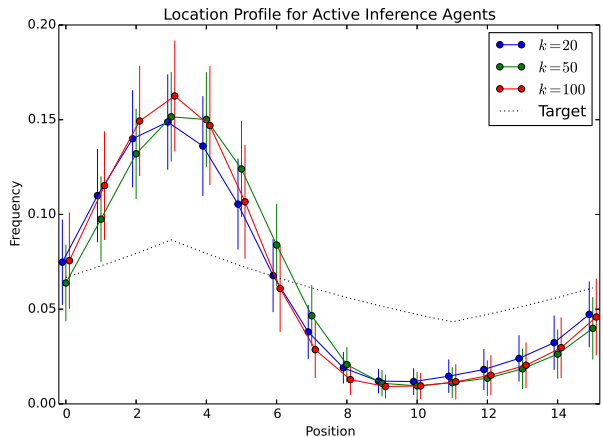}
\caption{Position frequencies over 500 time steps for agents updating their internal state using 20, 50 and 100 gradient descent iterations respectively. Curves are horizontally offset slightly for visual clarity. Points represent means of 300 runs; vertical bars show plus or minus one standard deviation. }
\label{fig:locationProfile}
\end{figure}

Although the agent's quality of inference falls noticeably with poorer IFE optimisation, it is not a priori obvious what effect this should have on its task performance. Figure \ref{fig:locationProfile}, shows the typical location profile (over 500 time steps) for agents using weak, moderate and strong optimisation procedures (characterised by 20, 50 or 100 gradient descent iterations between time steps). The agents were directed to maintain their location around a particular target and initialised in uniformly random locations. 

In fact, the statistical behaviour of a weakly-minimising agent appears remarkably similar to that of a strongly-minimising agent. This is interesting, because it suggests that accurate localisation is not particularly important for optimal control of position on this simple task. According to embodied / situated theories of cognition, many tasks can be accomplished by exploiting simple sensorimotor correlations directly, without the necessity for a high level of information processing; this simulation demonstrates the phenomenon clearly. Remember, the agent (indirectly, but provably) always selected the action on the basis of its (theorist-attributed) Bayesian belief. 

None of the agents' statistical location profiles over time closely approximate the target profile. This is an artefact of the simulation model, in which an agent's motor action is chosen deterministically to minimise IFE with respect to its target belief. Consequently, since it does not use any forward planning, the agent invariably attempts to move in the direction of its highest preference, even when preferences are closely matched. This could be easily rectified by including probabilistic motor control, where the agent's motor commands are real numbers determining the probabilities of particular actions; under such a scheme, motor commands must be minimised using a numerical optimisation scheme like other variables.


\section{Discussion}

The IFE principle  promises a exciting new account of action and perception within the same powerful framework. However the influence of these ideas has been hindered by the complexity of current formalism. Furthermore while central strength of the original formalism is it potential to unify both action and perception within a single theoretical framework yet to date agent based approaches have largely been absence except see \cite{FristonSM12, friston:reinf}.

While this claim may have some heuristic merit, it is worth noting that it is not without issues.
\begin{itemize}
	\item {Friston’s mathematical arguments depend on the assumption that sensory statistics are ergodic (i.e. that it makes sense to conflate the statistics of a system over a long time interval with its ``typical’’ behaviour at any instant); this assumption seems problematic in a biological context.}
	\item{Exporting thermodynamic entropy does not logically necessitate minimising sensory entropy or internal system state entropy (either thermodynamic or information-theoretic). It is entirely possible for more complex systems to be more stable (and better at self-regulation) than simpler systems.}
    

\end{itemize}

An ergodic system is one whose time average is the same as its state average; ergodicity is a physicist’s “trick” for making the maths of statistical mechanics more tractable. However, it is pretty clear that most biological systems are anything but ergodic in their dynamics. A human being is unlikely to experience the identically same waking sensory state twice in their lifetime. Certain primitive aspects of sensory state may be approximately ergodic (for instance, sensed core body temperature), but even there it seems plausible that there is scope for significant non-ergodicity.

Friston’s argument about minimising entropy is very similar to the argument presented in ``Every Good Regulator of a System Must Be a Model of That System'', \cite{ConAsh:regulator}. However, the authors of that paper explicitly concede that entropy is not a universally sound measure of successful regulation:
\begin{quote}
[Entropy and RMS error] tend to be similar... though the mathematician can devise examples to show that they are essentially independent.
\end{quote}

It is perhaps unfortunate that this ergodic-entropic reasoning has been presented as the primary motivation for the active inference framework, since the framework is consequently put in tension with the concept of information acquisition as an intrinsic motivation [dark room]. This tension comes from the mathematical fact that minimising surprisal also minimises information gain. 

There is no reason in principle why an agent proceeding according to active inference should not have expectations regarding its information gain as well as its sensory input, and recent research by Friston and colleagues \cite{Friston12:saccades}  explicitly models infotaxis by exactly such a method. However, these kinds of expectations regarding higher-order temporal features of internal dynamics do not fit well with the ergodic-entropic argument, which considers only ``detemporalised'' statistical features of external interactions with the environment. 

Moreover, the active inference framework can be motivated by more abstract considerations relating to

\subsection{Learning Without Reinforcement: Where Does The Intelligence Come From?}
One of the more disappointing aspects of the active inference framework is that it almost entirely fails to provide a mechanistic explanation for any specific pattern of intelligent behaviour. Under active inference, the agent’s sensorimotor behaviour emerges from the pre-existing structure of its sensory expectances. All of the agent’s intelligence is encoded in the mechanism which produces these expectances, about which the active inference framework itself has little or nothing to say.  

For machine learning or robotics applications, this is not a viable alternative to conventional paradigms such as reinforcement learning or supervised learning. The challenges faced in robot control lie precisely in understanding how particular desired external behaviours translate into sensory and motor patterns; in order to build a robot using active inference principles, it is necessary to specify exactly what the robot should expect to experience. 

It seems likely that the active inference framework is powerful enough that a reinforcement learning task could be re-formulated as an inbuilt expectance to receive reward signals, along with an internal model whose parameters can be modified to learn the relation between reward, action, sensation and environmental state. However, this merely transfers the problem to the question of how to implement such a model. It also raises the question of whether the active inference framework is too explanatorily powerful; in other words, is the “free-energy principle” falsifiable?
\nocite{*}
\printbibliography

\part*{Appendix}
\appendix

\section{Background}
\subsection{Kronecker delta}
\label{appx:kronecker}
Named after Leopold Kronecker, the Kronecker delta is a function of two variables, usually represented as $\delta_{x y}$:
\begin{equation}
\delta_{x y} =
\begin{cases} 
	1	& \mbox{ if } x = y \\
	0	& \mbox{ otherwise}
\end{cases}
\end{equation}

\subsection{Kullback-Leibler (KL) divergence}
\label{appx:kldiv}

Used as a measure of the difference between two probability densities (e.g. $q(\boldsymbol{x})$ and $p(\boldsymbol{x})$), the Kullback-Leibler divergence is defined by Solomon Kullback and Richard Leibler in \cite{kullback:kl} as:
\begin{equation}
	D(q(\mathbf{x}) \mid \mid p(\mathbf{x})) = \int q(\mathbf{x}) \log \frac{q(\mathbf{x})}{p(\mathbf{x})} \, d \mathbf{x}
\end{equation}

Important properties:
\begin{itemize}
	\item $D(q(\mathbf{x}) \mid \mid p(\mathbf{x})) \ne D(p(\mathbf{x}) \mid \mid q(\mathbf{x}))$ (the divergence is not symmetric)
	\item $D(q(\mathbf{x}) \mid \mid p(\mathbf{x})) \ge 0$
    \item $D(q(\mathbf{x}) \mid \mid p(\mathbf{x}) = 0 \iff q(\mathbf{x}) = p(\mathbf{x})$
\end{itemize}

KL divergence is measured in bits (if $\log_2 x$), bans (if $\log_{10} x$) or nats (if $\log_e x$), with the latters easily convertible to bits, and it can be considered as a quantity which gives the number of extra bits (nats or bans) required by the density $q(\mathbf{x})$ to represent the density $p(\mathbf{x})$.

\section{Calculating IFE and its Gradient}
\subsection{IFE}

We define the IFE $F$ as
\[ F(b', b, s, a) = \sum_{\psi'} P(\psi' \mid b') \log \frac{P(\psi' \mid b')}{P(\psi', s \mid b, a)} \]
\[ = - \mathcal{E}(b', b, s, a) - H(\Psi' \mid b') \]
\[ \mathcal{E}(b', b, s, a) = \sum_{\psi'}  P(\psi' \mid b') \log P(\psi', s \mid b, a) \]
\[ P(\psi', s \mid b, a) = \sum_{\psi} P(\psi' \mid \psi, a) P(s \mid \psi) P(\psi \mid b) \]
We can define $\text{Pre}_a(\psi')$ as the set of states which $\psi'$ can be reached from by performing action $a$, i.e. the set $\{\psi : P(\psi' \mid \psi, a) > 0 \}$, which will usually be smaller than the entire support of $\Psi$. Hence, we can rewrite $\mathcal{E}$ as 
\[ \mathcal{E}(b', b, s, a) = \sum_{\psi'} P(\psi' \mid b') \log \left( {\displaystyle \sum_{\psi \in \text{Pre}_a(\psi')} P(\psi' \mid \psi, a) P(s \mid \psi) P(\psi \mid b)} \right) \]

\subsection{Partial Derivatives}
With the free energy $F$ defined as
\[
	F(b', b, s, a) = \sum_{\psi'} q(\psi' \mid b') \ln \frac{q(\psi' \mid b')}{p(\psi', s \mid b, a)}
\]

the partial derivatives for each possible belief $b'_j$ are
\begin{equation}
	\frac{\partial F(b', b, s, a)}{\partial b'_j} = \sum_{\psi ' \in X} \frac{\partial q(\psi' \mid b')}{\partial b'_j} \left( 1 + \ln \frac{q(\psi' \mid b')}{p(\psi', s \mid b, a)} \right)
\end{equation}
where
\[ \frac{\partial q(\psi' \mid b')}{\partial b'_j} = q(\psi' \mid b') (\delta_{\psi' j} - q(j \mid b'_j))\]

with $\delta_{\psi' j}$ as the Kronecker delta described in \ref{appx:kronecker}.
\end{document}